\documentclass{article} % For LaTeX2e
\usepackage[T1]{fontenc}
\usepackage{iclr2027_conference,times}

\usepackage{amsmath,amsfonts,bm}

\def\eqref#1{equation~\ref{#1}}
\def\1{\bm{1}}

\DeclareMathAlphabet{\mathsfit}{\encodingdefault}{\sfdefault}{m}{sl}
\SetMathAlphabet{\mathsfit}{bold}{\encodingdefault}{\sfdefault}{bx}{n}

\usepackage{hyperref}
\usepackage{url}
\usepackage{wrapfig}
\usepackage{amsmath}
\usepackage{microtype}
\usepackage{graphicx}
\usepackage{adjustbox}
\usepackage{subcaption}
\usepackage{booktabs} % for professional tables
\usepackage{colortbl}
\usepackage{xcolor}
\usepackage{float}
\definecolor{tablehighlight}{RGB}{186, 186, 233}
\usepackage{multirow}
\usepackage{threeparttable}
\usepackage{algorithm}
\usepackage{algorithmic}
\usepackage{enumitem}

\usepackage{tikz}
\newcommand*\circled[1]{\tikz[baseline=(char.base)]{
            \node[shape=circle,fill,inner sep=1pt] (char) {\footnotesize \textcolor{white}{#1}};}}

\title{Imagine the Future, Internalize the Gist: Efficient VLA Reasoning via Internalized Spatiotemporal Imagination}

\author{
\textbf{Shenglan Li$^{1,*}$, Zhendong Mi$^{1,*,\ddagger}$, Hengyi Zhu$^{1,*}$, Jingwu Luo$^{2}$, Chun Kit Chan$^{3}$,}\\
\textbf{Geng Yuan$^{4}$, Yanzhi Wang$^{2}$, Pu Zhao$^{2}$, Shaoyi Huang$^{1,\dagger}$}\\[0.35em]
\normalfont\small
$^{1}$Stevens Institute of Technology \quad
$^{2}$Northeastern University\\
$^{3}$Arizona State University \quad
$^{4}$University of Georgia\\[0.25em]
$^{*}$Equal contribution \quad
$^{\ddagger}$Project lead \quad
$^{\dagger}$Corresponding author\\[0.2em]
\texttt{\{sli155,zmi2,hzhu37,shuang59\}@stevens.edu}\\
\texttt{\{luo.jingw,yanzhiwang,p.zhao\}@northeastern.edu}\\
\texttt{cchan44@asu.edu} \quad \texttt{geng.yuan@uga.edu}
}

\iclrfinalcopy
\begin{document}

\maketitle

% Replace the conference-status banner with the paper title, matching the
% clean running-header style used for the arXiv version.
\lhead{\scriptsize Imagine the Future, Internalize the Gist: Efficient VLA Reasoning via Internalized Spatiotemporal Imagination}
\hypersetup{
    colorlinks=false,
    pdfborder={0 0 1},
    citebordercolor={0.5294 0.8078 0.9216},
    linkbordercolor={0.5294 0.8078 0.9216},
    urlbordercolor={0.5294 0.8078 0.9216},
    pdftitle={Imagine the Future, Internalize the Gist: Efficient VLA Reasoning via Internalized Spatiotemporal Imagination},
    pdfauthor={Shenglan Li, Zhendong Mi, Hengyi Zhu, Jingwu Luo, Chun Kit Chan, Geng Yuan, Yanzhi Wang, Pu Zhao, Shaoyi Huang}
}

\begin{abstract}
Vision-language-action (VLA) models increasingly incorporate intermediate reasoning to improve robotic manipulation, yet existing approaches primarily reason about observed states without explicitly anticipating future scene evolution. Extending such reasoning to explicit future rollouts at every inference step, however, introduces substantial computational overhead. We propose \textbf{IG-VLA}, a VLA reasoning framework that enables models to \emph{\textbf{i}magine the future and internalize the \textbf{g}ist}. Our \textbf{Latent Spatiotemporal Reasoning} learns to imagine task-relevant future scene evolution directly in visual representation space, guiding action prediction without costly pixel-level video generation. To further reduce inference overhead, we introduce \textbf{Scene Gist Memory}, which internalizes reasoning-derived scene--behavior associations into a compact Scene Gist Token, preserving the benefits of future reasoning while bypassing explicit future imagination at inference. Extensive experiments on LIBERO, LIBERO-Plus, and VLABench demonstrate the effectiveness and efficiency of IG-VLA. On the LIBERO-Plus Language suite, both the reasoning and gist policies outperform the strongest baseline by nearly 6\% in success rate. The gist policy also achieves up to $6.38\times$ speedup over baselines, reducing inference latency from 1,081\,ms to 169.5\,ms per action chunk on a single NVIDIA A6000 GPU. These results demonstrate that future spatiotemporal reasoning can be effectively internalized for efficient VLA deployment.

% Vision-language-action (VLA) models increasingly incorporate intermediate reasoning to improve robotic manipulation, yet existing approaches mainly reason over observed states without explicitly anticipating how the environment may evolve.
% Moreover, performing explicit reasoning rollouts at every inference step introduces substantial computational overhead.
% In this work, we propose \textbf{IG-VLA}, a VLA reasoning framework that enables models to \emph{\textbf{i}magine the future and internalize the \textbf{g}ist}.
% Specifically, we introduce \textbf{Latent Spatiotemporal Reasoning}, which learns to imagine task-relevant future scene evolution directly in the visual representation space, providing anticipated spatial and environmental dynamics for action prediction without costly pixel-level video generation.
% We further introduce \textbf{Scene Gist Memory}, which internalizes reasoning-derived scene--behavior associations into a compact Scene Gist Token, allowing the model to preserve the benefits of future reasoning while bypassing explicit future imagination at inference.
% Extensive experiments on LIBERO, LIBERO-Plus, and VLABench demonstrate the effectiveness and efficiency of our framework.
% On LIBERO-Plus Language suite, IG-VLA with either reasoning or gist policy outperform the strongest baseline by nearly 6\% in success rate.
% Meanwhile, the gist policy reduces inference latency from 263\,ms of ImageWAM baseline to 169.5\,ms on a NVIDIA A6000 GPU, demonstrating that future spatiotemporal reasoning can be effectively internalized for efficient VLA deployment.
\end{abstract}

\section{Introduction}

%\textcolor{red}{abstract + intro, 2 pages}

% Vision-Language-Action (VLA) models have  emerged as a powerful framework for general-purpose robotic manipulation, enabling robots to translate multimodal observations and language instructions into executable actions
Vision-Language-Action (VLA) models enable general-purpose robotic manipulation by mapping multimodal observations and language instructions to actions
\cite{guan2025efficient,li2025vla,li2025comprehensive,sapkota2025vision,wang2025vision}. 
% However, learning an effective end-to-end mapping from multimodal perception to robotic actions remains highly demanding, largely due to its dependence on extensive, high-quality demonstration data 
However, learning an effective end-to-end mapping from multimodal perception to robotic actions requires extensive, high-quality demonstration data
\cite{liu2025aligning,ma2026survey}. 
% To alleviate this difficulty, recent VLA models increasingly adopt a “think-before-acting” paradigm through Chain-of-Thought (CoT) reasoning 
To address this challenge, recent VLA models adopt a ``think-before-acting'' paradigm through Chain-of-Thought (CoT) reasoning
\cite{guo2025vla,ye2025vla,yin2025deepthinkvla}.By reasoning over task goals, visual observations, and object relations before execution, these models decompose complex instructions and adapt action plans to environmental feedback.

\begin{figure*}[t]
    \centering
    % \vspace{-0.1in}
    \includegraphics[
        width=0.95\textwidth,
        height=0.72\textheight,
        keepaspectratio
    ]{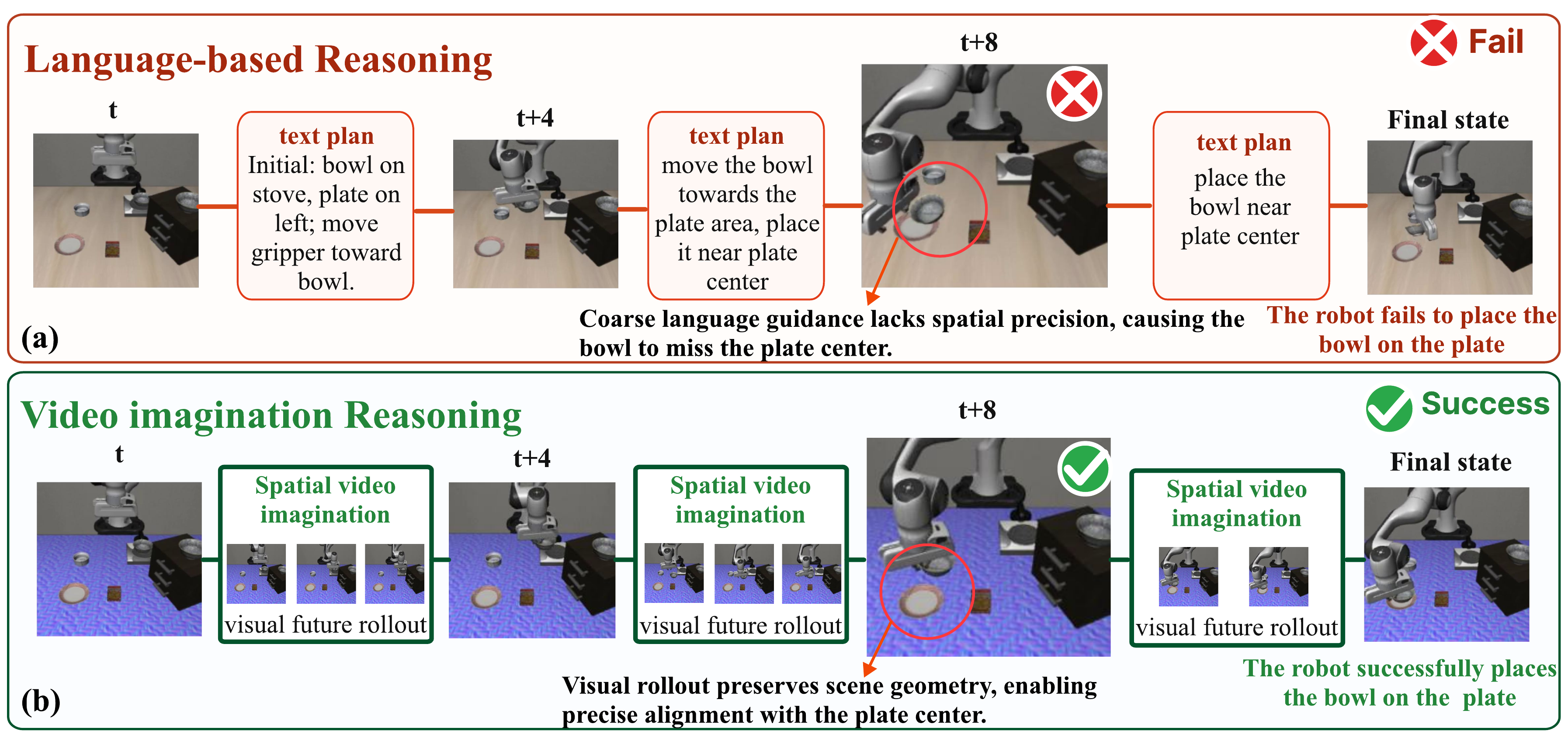}
    \vspace{-0.1in}
    \caption{
    Language-based vs. latent visual reasoning.
    \textbf{(a)} Language-based reasoning may lose fine-grained spatial information, leading to imprecise actions;
    \textbf{(b)} 
    % Latent visual reasoning preserves spatial details while anticipating scene evolution, enabling more precise actions.
    Latent visual reasoning preserves evolving scene geometry through visual future reasoning, enabling more precise actions.
    }
    \vspace{-0.2in}
    \label{fig:compare}
\end{figure*}

Existing VLA paradigms
% reasoning paradigms for VLA models 
include \textit{language-based reasoning} \cite{zawalski2024robotic,lee2025molmoact,qu2508eo,huang2026fast}, which carry out reasoning in  language domain and \textit{reasoning in vision},
exemplified by CoT-VLA's decomposition of predicted goals into visual subgoals \cite{zhao2025cotvla}
and 
% such as CoT-VLA \cite{zhao2025cotvla} to decompose a predicted goal into a sequence of visual subgoals, or 
VLA-Thinker's repeated incorporation of intermediate state images~\cite{wang2026vla}.
% to repeatedly inject   in reasoning.
As shown in Figure \ref{fig:compare}(a), language-based reasoning loses fine-grained visual details and achieves lower success rates.
% language based reasoning discards fine-grained visual details with relatively lower success rate. 
% In contrast, reasoning in vision  methods can mitigate the gradual drift from the actual environment with updated visual information as it unfolds
In contrast, incorporating updated visual information during reasoning can mitigate drift from the actual environment
\cite{wang2026vla} (Figure \ref{fig:compare}(b)).

% Existing reasoning paradigms for VLA models can be broadly categorized into two types. The first is \textit{language-based reasoning} \cite{zawalski2024robotic,lee2025molmoact,qu2508eo,huang2026fast}: visual observations are encoded into fixed representations at the beginning of reasoning and remain unchanged thereafter. As a result, all subsequent reasoning process is carried out largely in the language domain.
% the reasoning process is largely carried out in the language domain.
%This can cause reasoning to gradually drift from the actual environment as it unfolds \cite{wang2026vla}, and it discards fine-grained visual detail that later becomes relevant to the decision, as shown in Figure \ref{fig:compare}(a).

% However, language-centric reasoning can gradually detach the model from the actual environment as the reasoning process unfolds \cite{wang2026vla}, while also overlooking fine-grained visual details that are critical for effective decision-making, \textcolor{red}{as shown in Figure \ref{}.}
% The second paradigm instead \textit{re-grounds reasoning in vision}:
% % In contrast, the second paradigm continuously incorporates visual information into the reasoning process, grounding intermediate decisions in the evolving environment. 
% for example, CoT-VLA \cite{zhao2025cotvla} decomposes a predicted goal into a sequence of visual subgoals, while VLA-Thinker \cite{wang2026vla} repeatedly injects intermediate state images to provide updated visual context and to keep reasoning anchored to the current scene. This addresses the drift problem of the first paradigm, but it leaves a different gap:  These so-called visual reasoning methods still largely rely on currently observed visual inputs to guide subsequent actions. The visual information mainly informs the model of what has already been accomplished and the final goal. However, they still do not explicitly leverage visual reasoning about how the environment will evolve and how actions should be planned and done accordingly.

In general, existing approaches face two limitations: (i) reliance on observed states rather than future predictions, and (ii)
substantial computational overhead from repeated reasoning rollouts.
% substantial computation overhead for full spatiotemporal reasoning rollouts at every inference step. 
Specifically, existing approaches primarily reason over observed states and plan subsequent actions based on the current environment, without explicitly anticipating how the environment may evolve under different decisions. Meanwhile, 
reasoning in VLA models need to perform full spatiotemporal reasoning rollouts repeatedly at every inference step, introducing substantial computation overhead. 

To address these issues, we propose a visual reasoning framework for VLA models with two core mechanisms: \textbf{latent spatiotemporal reasoning} and \textbf{scene gist memory}. 
Latent spatiotemporal reasoning 
enables the model to imagine future scene evolution in visual representation space, exploring plausible action trajectories and their potential outcomes.
% 
% reason through imagined images or videos of future states.
% By simulating scene evolution in the form of spatial videos, the model explores plausible action trajectories and their potential outcomes under changing environments. 
% 
These imagined futures guide action selection, as shown in Figure \ref{fig:compare}(b).
% The VLA model can simulate future scene evolution in the form of spatial videos, imagining plausible action trajectories and their potential outcomes under changing environmental states. Based on imagined futures and evolutions, it reasons and determines appropriate actions, as shown in Figure \ref{fig:compare}(b).  
% 
It is inspired by  human reasoning process,  which imagines how a scene might unfold under different candidate actions and makes decisions based on  expected consequences \cite{atance2001episodic}. 
For example, when crossing a street,  we consider multiple possible futures conditioned on our own actions such as crossing, waiting, or speeding up, then choose the safest action.
%choose the action with the safest expected outcome.
This anticipation
% is \textit{action-conditioned}: it 
evaluates candidate behaviors through the imagined consequences rather than merely extrapolating scene evolution independently of  actions \cite{conti2021harnessing}.

\begin{figure*}[t]
    \centering
    \includegraphics[width=1 \textwidth]{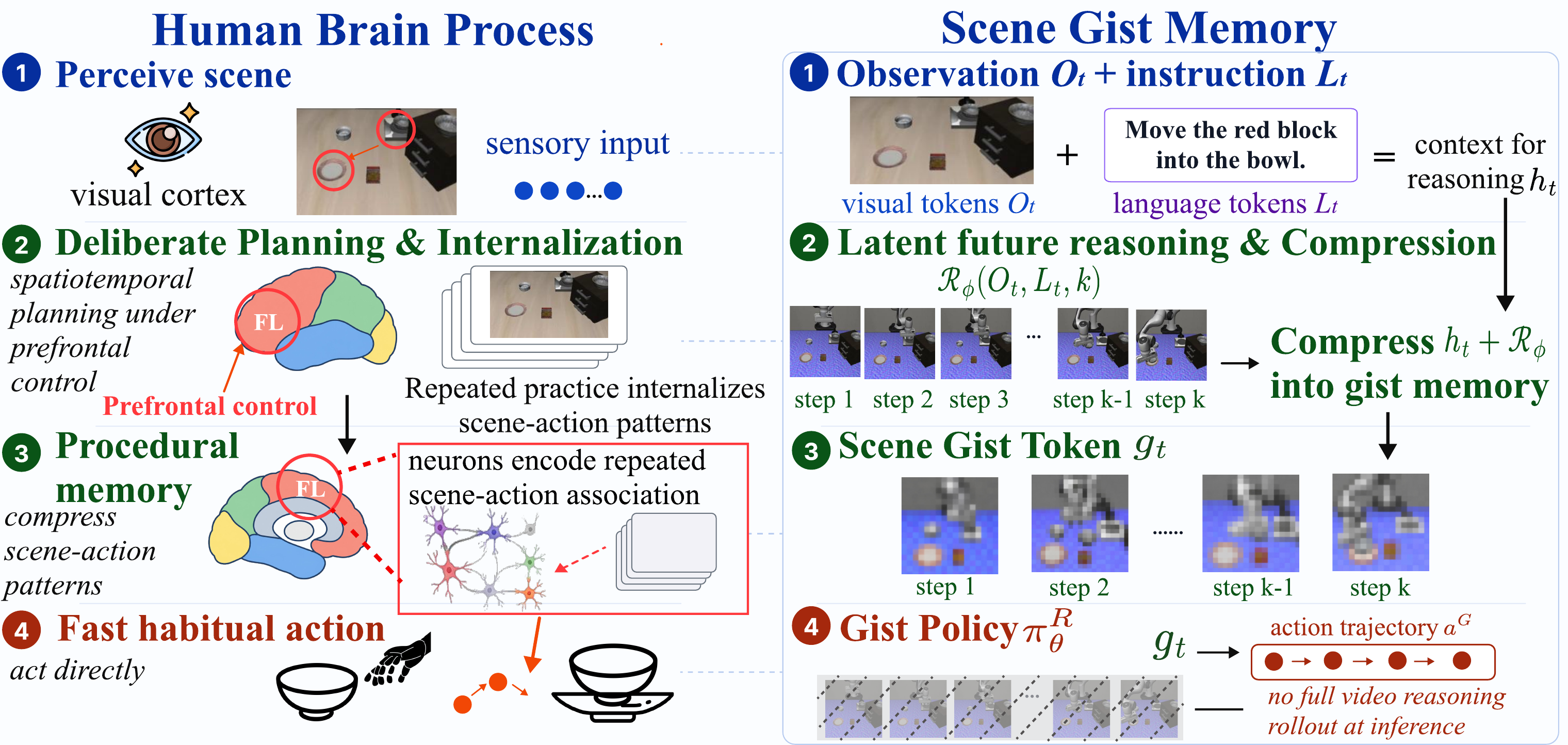}
    \vspace{-0.2in}
    \caption{Human Automaticity vs. Scene Gist Memory.}
    \vspace{-0.1in}
    \label{fig:brain}
\end{figure*}

To reduce computational overhead, our proposed Scene Gist Memory internalizes reasoning-derived scene--behavior associations into a compact \textit{Scene Gist Token}. This token captures general scene context and corresponding action patterns rather than preserving the complete future reasoning trajectory. During inference, the gist encoder predicts the Scene Gist directly from the current context, allowing the model to bypass the explicit future-reasoning pathway while retaining reasoning-derived behavioral guidance (Figure \ref{fig:brain}).
% achieving efficient decision making with fewer model  neurons 
% 
It's inspired
% It is motivated 
by human behavioral psychology that repeated 
experience
% reasoning and actions 
in familiar environments 
helps consolidate learned behaviors into procedural memory involving the basal ganglia~\cite{wulf2010motor,wood2016psychology}.
This memory supports rapid responses with less deliberation, reducing cognitive effort and decision time.
% can gradually become automatized in human's brain: coarse associations between familiar scenes and corresponding behaviors are compressed into basal ganglia and consolidated into procedural memory, allowing humans to respond rapidly to similar situations without the  need for extensive deliberative processing. 
% This substantially reduces the cognitive effort and time for decision making. 
For example, familiarity with a home allows a person to head directly to the kitchen for coffee without searching the entire house. Repeated experience reinforces the association between the familiar scene and the corresponding navigation behavior.
Our main contributions are summarized as follows:
\vspace{-0.1in}
\begin{itemize}
% [leftmargin=*, noitemsep, topsep=0pt]
    \item Our framework incorporates \textbf{Latent Spatiotemporal Reasoning}, enabling VLA models to learn to imagine task-relevant scene evolution in visual representation space. These anticipated scene dynamics guide action prediction without costly pixel-level video generation.   
    % Our framework incorporates \textbf{Latent Spatiotemporal Reasoning}, a future-oriented reasoning paradigm that enables VLA models to learn to imagine task-relevant scene evolution directly in the visual representation space. By anticipating potential environmental and spatial changes before acting, the model can leverage future scene dynamics to better reason about and predict appropriate actions without costly pixel-level video generation.

    \item We further introduce \textbf{Scene Gist Memory}, which internalizes reasoning-derived scene--behavior associations into a compact Scene Gist Token. It enables the model to 
    reuse this knowledge in corresponding scenarios, preserving reasoning benefits while bypassing explicit future imagination and full reasoning rollouts at inference.
    % directly leverage previously learned behavioral knowledge in corresponding scenarios, preserving the reasoning benefits %of future reasoning 
    % while bypassing explicit future imagination and the full reasoning rollout at inference.

    \item Extensive experiments on LIBERO, LIBERO-Plus, and VLABench demonstrate the effectiveness and efficiency of our framework. 
For example, IG-VLA achieves an 85.6\% success rate on LIBERO-Plus's challenging Language suite, outperforming the strongest baseline by nearly 6\%. 
Meanwhile, the gist policy requires only 169.5\,ms per action chunk on a single NVIDIA A6000, achieving $6.38\times$, $1.78\times$ and $1.55\times$ speedups over FastWAM-IDM (1,081\,ms), FastWAM (1-step) (302\,ms) and ImageWAM (263\,ms), respectively.
% Meanwhile, the gist policy requires only 169.5\,ms per action chunk on a single NVIDIA A6000, substantially faster than FastWAM (302\,ms) and ImageWAM (263\,ms).
    % \textcolor{red}{Please double check the number.
    % }
    
    % future spatiotemporal reasoning can be effectively internalized for efficient VLA deployment.
\end{itemize}

\section{Related Work}
\vspace{-0.1in}
%\textcolor{red}{0.5 pages}

\textbf{VLA Models.}
Recent advances in embodied AI have shifted robot learning toward foundation VLA models that directly map visual observations and language instructions into executable actions \cite{brohan2022rt,mees2024octo,nvidia2025gr00t,yang2025magma,pertsch2025fast,driess2026knowledge}. OpenVLA \cite{kim2024openvla} adopts an end-to-end policy learning framework for language-conditioned robotic manipulation. $\pi_0$ \cite{black2024pi_0} extends this paradigm with scalable policy learning across more diverse robotic tasks. HAMSTER \cite{li2025hamster} strengthens spatial reasoning through structured visual representations,
while TraceVLA \cite{zheng2025tracevla} further incorporates 2D visual trajectories to enhance the alignment between perception and action. 
%Although these imitation-based foundation VLA models achieve strong performance on common manipulation tasks \cite{huang2026thinkact}, their reliance on finite demonstration datasets limits their scalability to complex environments and their ability to perform long-horizon planning.

\textbf{VLA Reasonings.}
%Reasoning has become a critical component for improving decision-making in large language models (LLMs), with CoT demonstrating remarkable gains across mathematical reasoning, code generation, and question answering \cite{wei2022chain,wang2024chain,yeo2025demystifying}. 
% CoT in LLMs demonstrates remarkable gains across mathematical reasoning, code generation, and question answering \cite{wei2022chain,yeo2025demystifying}. 
% Inspired by these advances, recent studies extend explicit reasoning to VLA models \cite{huang2026thinkact}. 
% 
CoT improves LLM performance in mathematical reasoning, code generation, and question answering \cite{wei2022chain,yeo2025demystifying}, motivating explicit reasoning in VLA models \cite{huang2026thinkact}.
Existing approaches mainly follow several directions. Prompting-based methods such as EcoT \cite{zawalski2024robotic} use supervised reasoning traces to decompose tasks into intermediate subgoals before action execution.
% , decompose complex tasks into intermediate subgoals through supervised reasoning traces before action execution. 
% 
Data-centric methods, including RAD \cite{clark2025action}, combine human videos with LLMs to construct reasoning-action demonstrations for policy learning.
% exploit large-scale human videos and standard LLMs to construct reasoning-action demonstrations for policy learning. 
Another line of work, represented by CoT-VLA \cite{zhao2025cot}, replaces textual reasoning with visual subgoal representations to better align planning with robot perception. Recently, reinforcement learning  \cite{shao2024deepseekmath,guo2025deepseek} has been adopted to optimize reasoning policies under verifiable rewards, where  GRPO encourages more effective reasoning trajectories % for vision-language tasks
\cite{feng2026video,tan2025reason,liu2025visual}. However, current methods often require costly reasoning supervision and sophisticated reward engineering \cite{huang2026fast}. 

%\begin{figure*}[t]
%    \centering
%    \includegraphics[width=0.5 \textwidth]{fig/text_video.pdf}
%    \caption{language.}
%    \label{fig:language}
%\end{figure*}

%\input{sections/method_old}

\vspace{-0.1in}
\section{Methodology}
\vspace{-0.1in}

% \textcolor{red}{method 3.5 pages}

\begin{figure*}[t]
    \centering
    \includegraphics[width=1\textwidth]{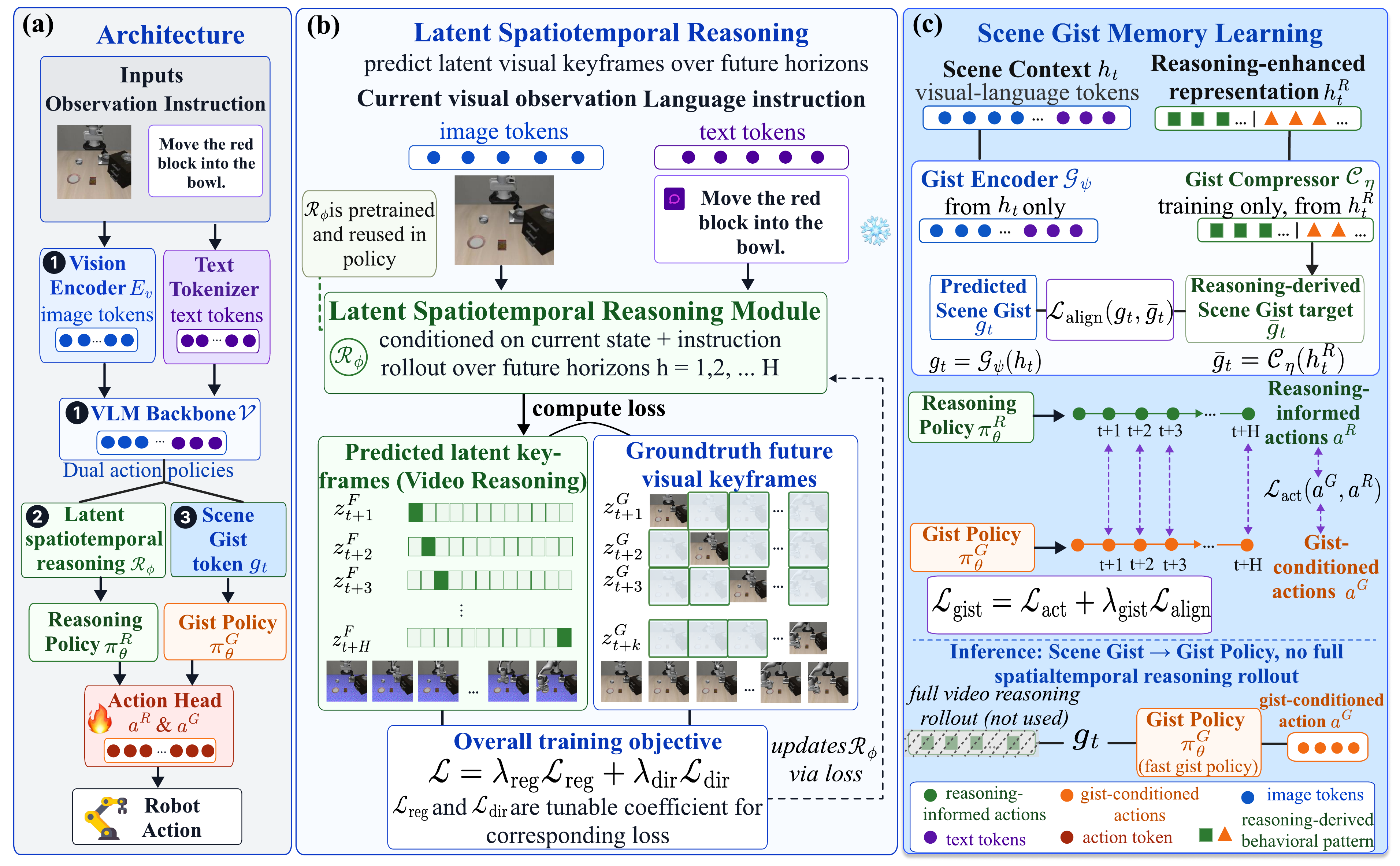}
    \vspace{-0.2in}
    \caption{
    Overview of our framework.
    \textbf{(a)} the overall architecture; 
    % supports our reasoning policy and the lightweight gist policy.
    \textbf{(b)} the latent spatiotemporal reasoning module learns to predict future visual keyframes;
    \textbf{(c)} the gist modules internalize reasoning-derived behavioral information into a compact Scene Gist Token.
    }
    \vspace{-0.1in}
    \label{fig:framework}
\end{figure*}

In this section, we present IG-VLA, a VLA framework that incorporates two core mechanisms, reasoning over possible future predicted trajectories to improve robustness and task success in long-horizon embodied tasks, and scene gist token to capture a coarse-grained scene-behavioral impression for efficient inference, as shown in Figure \ref{fig:framework}.

%In this section, we present XXX, a VLA framework that leverages the current environmental state to perform spatiotemporal reasoning over possible future trajectories, thereby improving robustness and task success in long-horizon embodied tasks and then we distill the actions of the video-reasoning model into a lightweight policy to enable efficient on-device deployment, allowing it to directly respond to familiar states without repeatedly performing costly spatiotemporal reasoning. 
%Our method is built upon following components: First, in Section 3.2 we augment the VLA model with a learnable spatiotemporal reasoning module that anticipates future visual trajectories conditioned on the current environmental state. The resulting video-based reasoning is internalized into the model, enabling it to reason over future scene dynamics without explicitly generating intermediate videos during inference. Second, in Section 3.3 we distill the action behaviors of the teacher video-reasoning model into a lightweight student policy, enabling the student to learn how the teacher responds to different environmental states.

\vspace{-0.1in}
\subsection{Model Architecture}
% \vspace{-0.1in}

We first introduce our overall model architecture, % of our VLA model, 
as illustrated 
in Figure~\ref{fig:framework}. Our model builds upon a conventional VLA architecture, and incorporates  spatiotemporal reasoning modules and gist modules.
%consisting of a vision encoder, a vision-language backbone, and an action head, while introducing a latent Video-CoT module for spatiotemporal reasoning. 

% Given the current visual observation $o_t$ and language instruction $L_t$, 
% the model first extracts multimodal representations of the current environment. 
% The latent Video-CoT module then reasons about possible future scene evolution 
% in the representation space, and the resulting reasoning-enhanced features are 
% used by the action head to generate robot actions.
\circled{1} \textbf{Vision Encoder and Vision-Language Backbone.}
A pretrained vision encoder $E_v$ maps the current visual observation $o_t$ to visual tokens:
% \begin{equation}
% e_t=E_v(o_t).
% \end{equation}
$e_t=E_v(o_t)$.
These tokens compactly encode scene semantics and spatial structure, supporting subsequent spatiotemporal reasoning.
% instead of operating directly on raw pixels.
% Given the current visual observation $o_t$, we employ a pretrained vision  encoder $E_v$ to extract visual representations:
% \begin{equation}
%     e_t=E_v(o_t),
% \end{equation}
% where $e_t$ denotes the visual tokens encoding the current scene.  Compared with directly operating on raw pixels, these representations provide  compact semantic and spatial information for subsequent spatiotemporal  reasoning.
% 
%\textbf{Vision-Language Backbone.}
% The language instruction  is tokenized into textual tokens  $L_t$. The visual and textual tokens are   jointly processed by a  pretrained vision-language backbone $\mathcal{V}$:
A pretrained vision-language backbone $\mathcal{V}$ jointly processes $e_t$ and the tokenized language instruction $L_t$:
% \begin{equation}
%     h_t
%     =
%     \mathcal{V}
%     \left(
%         e_t,L_t
%     \right),
% \end{equation}
$ h_t=\mathcal{V}\left( e_t,L_t\right)$, 
where $h_t$ encodes the current observation and task instruction, providing multimodal context for future-scene reasoning and action generation.
% represents the multimodal hidden representation of the current  observation and task instruction. It provides   contextual  information   for both future-scene reasoning and action generation.

\circled{2} \textbf{Latent Spatiotemporal Reasoning Module and Reasoning Policy.}
We introduce a latent spatiotemporal reasoning module $\mathcal{R}_{\phi}$ to anticipate task-relevant scene evolution and guide action trajectory selection. Rather than autoregressively generating pixel-level frames, $\mathcal{R}_{\phi}$ predicts future states directly in visual representation space:
%\textcolor{red}{(give a formal name for all modules, and when mention the modules, use the formal names. there are many names for the same thing, such as latent spatiotemporal reasoning module, spatiotemporal reasoning module, and latent future reasoning)}
% To explicitly reason about how the current environment may evolve and what trajectory to choose for tasks under the given instruction, we introduce a latent spatiotemporal reasoning module  $\mathcal{R}_{\phi}$. Instead of autoregressively generating future frames in  pixel space, $\mathcal{R}_{\phi}$ predicts future scene evolution directly in  the visual representation space: 
%\textcolor{red}{(replace $o_t$ with $h_t$ or $e_t$ in following equations?)}
\begin{equation}
    z_{t+k}^\mathcal{F}
    =
    \mathcal{R}_{\phi}
    \left(
        h_t,L_t,k
    \right),
    \qquad
    k=1,\ldots,H,
\end{equation}
where $z_{t+k}^\mathcal{F}$ denotes the predicted scene representation at the subsequent $k$-th future  step. Together, these latent states form a compact trajectory of anticipated spatiotemporal dynamics. %, supporting action planning without reconstructing video frames.

% denotes the predicted latent representation of the scene  at the subsequent $k$-th future  step.  These future latent states collectively form a compact trajectory that captures the anticipated spatiotemporal evolution of the environment, allowing the model to reason about task-relevant future dynamics and select an appropriate action trajectory without explicitly reconstructing video frames.

Our framework consists of a reasoning policy that %leverages the predicted latent future trajectory for deliberative action prediction. %, and  a lightweight gist policy that directly predicts actions from the compressed  scene gist token.
%For the reasoning policy, 
%The action head 
predicts the future actions based on the current multimodal representation  and the predicted   latent future representation:
% Our framework comprises a reasoning policy that uses the predicted latent future trajectory for deliberative action prediction and a lightweight gist policy that predicts actions directly from a compressed Scene Gist Token. For the reasoning policy, the action head combines the current multimodal context with the predicted future representation:
\begin{equation}
a^{R}_{t+k}
=
\pi^{R}_{\theta}
\left(
h_t,L_t,z^{F}_{t+k}
\right).
\end{equation}
where $a^{R}_{t+k}$ denotes  predicted action   at the $k^{th}$ future step. It explicitly exploits anticipated scene evolution to determine future actions and provides reasoning-informed action guidance in training.

\circled{3} \textbf{Gist Compressor, Gist Encoder, and Gist Policy.} 
We introduce a gist compressor and a gist encoder to summarize reasoning-derived information and predict it directly from the current context, respectively. After the complete latent spatiotemporal reasoning rollout, the implicit action reasoner fuses the current visual--language context with all predicted future representations to obtain a reasoning-enhanced representation $h_t^R$. The gist compressor $\mathcal{C}_{\eta}$ maps this representation through an MLP bottleneck into a single Scene Gist vector:
\begin{equation}
    \bar g_t = \mathcal{C}_{\eta}(h_t^R).
\end{equation}
The compressor is trained by matching its induced actions to those of the full reasoning policy, so that $\bar g_t$ retains behaviorally relevant information from the complete reasoning pathway and serves as a target for the gist encoder. The gist encoder $\mathcal{G}_{\psi}$ then learns to predict this target using only the current observation and instruction, represented by $h_t$:
% \textcolor{red}{(I am not sure if we need to mention gist compressor. it seems a bit complex. we use compressor to obtain scene gist vector, then use encoder to predict the vector and obtain gist token. we never discuss how to train the compressor. equation 4 is also different from equation 9.)}:
\begin{equation}
    g_t = \mathcal{G}_{\psi}(h_t).
\end{equation}
The encoder concatenates the masked means and standard
deviations of the current prefix key--value features across layers, normalizes the resulting statistics, and maps them to a gist vector through a two-layer MLP. The gist policy uses the predicted $g_t$ in place of the implicit action reasoner's output. The existing interaction module incorporates this vector into action generation:
\begin{equation}
    a^G_{t:t+H-1}
    = \pi^G_{\theta}(h_t,L_t,g_t).
\end{equation}
The original visual--language prefix and explicit action reasoning module are retained, allowing the policy to combine detailed visual information with the behavioral guidance provided by the gist. At inference, the model predicts the gist directly from the current context, bypassing the latent spatiotemporal reasoning rollout, the implicit action reasoner, and the gist compressor.

\subsection{Model Training with Future Reasoning and Scene Gist Memory}

Our VLA framework   incorporates two core mechanisms, reasoning over possible future predicted trajectories 
%(including reasoning module and reasoning policy)  
and scene gist tokens. %(including gist encoder and gist policy). 
We first train the reasoning policy and then train the Scene Gist modules with the pretrained reasoning policy frozen. The training details are specified below.

\subsubsection{Learning Spatiotemporal Future Reasoning For Trajectory Alignment}

%To enable VLA models to reason about future trajectories through spatiotemporal scene evolution, we introduce a learnable reasoning module into the VLA model which encode future visual observations into visual latent representations \cite{xie2026latentvla} to avoid the overhead of explicitly generating intermediate videos. This allows the model to perform spatiotemporal reasoning entirely in the latent space without explicit video decoding.

We first  encode the visual observation into a latent visual representation $z_t$. Instead of explicitly reconstructing future observations at  pixel level,  we train a spatiotemporal reasoning module $\mathcal{R}_{\phi}$ to anticipate  environment dynamics and potential trajectory directly in  visual representation space. Specifically, conditioned on the current visual state and task instruction, 
the reasoning module predicts
\begin{equation}
    z_{t+k}^\mathcal{F}
    = \mathcal{R}_{\phi}(h_t, L_t, k),
\end{equation}
and the anticipated future representation can be expressed as
% \begin{equation}
%     \hat{e}_{t+k} = e_t + \widehat{\Delta e}_{t,k}.
% \end{equation} 
$ \hat{z}_{t+k} = z_t + z_{t+k}^\mathcal{F}$.
Here, $z_{t+k}^\mathcal{F}$ represents the predicted visual transition from the current state at time $t$ to the $k^{th}$ future step, capturing how the scene is expected to evolve in the latent space. The ground truth is $  z_{t+k}^G = z_{t+k} - z_t$. 
%Accordingly, $\hat{e}_{t+k}$ denotes the predicted visual representation of the future state. 
Predicting the visual transition rather than the complete future representation encourages the reasoning module to focus on task-relevant changes in the environment, such as object displacement, robot motion, and interaction events, while reducing the need to reproduce static scene information already contained in the current observation, as shown in Figure \ref{fig:framework}.

We supervise the predicted future evolution using the ground truth visual transitions extracted from demonstration trajectories. The primary objective minimizes the discrepancy between them: %the predicted and ground truth future transitions:
%\vspace{-0.1in}
\begin{equation}
\small    \mathcal{L}_{\mathrm{reg}}
    =
    \frac{1}{K}
    \sum_{k=1}^{K}
    \left\|
        z_{t+k}^\mathcal{F} - z_{t+k}^G
    \right\|_2^2.
\end{equation}
\vspace{-0.1in}

With $\mathcal{L}_{\mathrm{reg}}$ encouraging   a magnitude similarity, 
%we further require the predicted trajectory to align with the ground truth in its direction of evolution. Therefore, 
an additional loss is adopted to encourage directional consistency between the predicted  and target transitions:
\begin{equation}
\small    \mathcal{L}_{\mathrm{dir}}
    =
    \frac{1}{K}
    \sum_{k=1}^{K}
    \left(
        1 -
        \cos\left(
            z_{t+k}^\mathcal{F},
            z_{t+k}^G
        \right)
    \right).
\end{equation}

\subsubsection{Internalizing Spatiotemporal Reasoning into Scene Gist Memory}
% Inspired by human behavioral automaticity, we seek to internalize repeatedly encountered scene--behavior associations into a compact memory representation. Humans gradually automatize frequently practiced behaviors: coarse-grained scene memory and their corresponding behaviors can be consolidated into procedural memory, enabling rapid responses without engaging extensive neural resources for deliberative reasoning.
% Motivated by this mechanism, while learning spatiotemporal future reasoning, we simultaneously compress the coarse scene context and its reasoning-derived behavioral pattern into a compact \textit{Scene Gist Token}. Rather than preserving the complete future reasoning trajectory, this token captures a coarse-grained scene-behavioral impression, i.e., general association between the scene and corresponding actions, which allows VLA model to inference with fewer parameters or a more lightweight policy. \textcolor{red}{This paragraph is mentioned in introduction.}
We train the Scene Gist modules in two stages.
First, the gist compressor $\mathcal{C}_{\eta}$ is trained with the reasoning policy frozen to compress the reasoning-enhanced representation $h_t^R$ into a behaviorally informative target $\bar g_t$.
Specifically, $\bar g_t$ replaces the implicit action reasoner's output, and the compressor is optimized by matching the resulting actions to those of the full reasoning policy.
We then freeze the compressor and train the gist encoder to predict $g_t=\mathcal{G}_{\psi}(h_t)$ directly from the current context.
The encoder is supervised by both representation alignment and action consistency:
\begin{equation}
\mathcal{L}_{\mathrm{align}}
=
\left\|
g_t-\operatorname{sg}(\bar g_t)
\right\|_2^2,
\qquad
\mathcal{L}_{\mathrm{act}}
=
\frac{1}{H}
\sum_{k=0}^{H-1}
\left\|
a^G_{t+k}
-
\operatorname{sg}(a^R_{t+k})
\right\|_2^2.
\end{equation}
Here, $\operatorname{sg}(\cdot)$ denotes stop-gradient.
$\mathcal{L}_{\mathrm{align}}$ encourages the predicted Scene Gist to match the reasoning-derived target produced by the frozen gist compressor, while $\mathcal{L}_{\mathrm{act}}$ encourages the gist policy to preserve the action behavior of the full reasoning policy.

The final Scene Gist objective is
\begin{equation}
\mathcal{L}_{\mathrm{gist}}
=
\mathcal{L}_{\mathrm{act}}
+
\lambda_{\mathrm{gist}}
\mathcal{L}_{\mathrm{align}},
\end{equation}
where $\lambda_{\mathrm{gist}}$ weights the representation alignment loss relative to the action consistency loss.

\textbf{Efficient Inference with Gist Token.}
% During training process, we need compress the behavioral knowledge acquired through  spatiotemporal reasoning into a compact \textit{Scene Gist Token}.  Rather than encoding the complete future reasoning trajectory, the token captures a coarse association between the current scene, task instruction, and the corresponding behavioral pattern. Consequently, 
At inference time, the gist encoder directly predicts a single Scene Gist $g_t$ from the current context and uses it to condition the existing action-generation pathway for the entire action chunk. The latent reasoning rollout, implicit action reasoner, and gist compressor are therefore bypassed, substantially reducing inference overhead.

\section{Experiments}

% \textcolor{red}{Experiments 3 pages}

\subsection{Experimental Setup}
\textbf{Dataset.} 
% We use LIBERO as the primary training benchmark. Models are trained on LIBERO demonstrations and evaluated on both LIBERO and zero-shot LIBERO-Plus. This setting tests whether latent visual reasoning improves standard manipulation performance and whether the learned reasoning representation transfers to perturbed environments without additional demonstrations.
% We additionally evaluate on VLABench, a language-conditioned tabletop manipulation benchmark with multiple generalization tracks. For controlled VLABench comparison, we follow the released ACoT-VLA VLABench setting and fine-tune both the No-Video-CoT baseline and our Video-CoT variant on the same VLABench training data. The two models are evaluated on the same public tracks: in-distribution, cross-category, commonsense, semantic-instruction, and unseen-texture.
We use LIBERO~\cite{liu2023libero} as the primary training benchmark and evaluate success rates on the Spatial, Object, Goal, and Long suites.
We further evaluate robustness on LIBERO-Plus~\cite{Fei_2026_CVPR} 
%under zero-shot transfer and 
supervised fine-tuning.
%, where the former directly transfers the LIBERO-trained policy to perturbed environments and the latter further and trains it on LIBERO-Plus.
Finally, we evaluate on VLABench \cite{zhang2025vlabench} following~\cite{zhong2026acotvla}, where the Non-Latent Spatiotemporal Reasoning baseline and our latent spatiotemporal reasoning module are fine-tuned on identical training data and evaluated on five public tracks: in-distribution, cross-category, commonsense, semantic-instruction, and unseen-texture. %\textcolor{red}{LIBERO-Plus supervised fine-tuning not mentioned.}

\noindent\textbf{Baselines.} 
% Our primary baseline is ACoT-VLA \cite{zhong2026acotvla}, which uses explicit and implicit action chain-of-thought reasoning for action generation. We also compare with representative VLA and world-model baselines reported under the corresponding benchmark protocols, including Octo~\cite{mees2024octo}, OpenVLA~\cite{kim2024openvla}, $\pi_0$~\cite{black2024pi_0}, $\pi_{0.5}$~\cite{physicalintelligence2025pi05}, CoT-VLA~\cite{zhao2025cotvla}, WorldVLA~\cite{cen2025worldvla}, DreamVLA~\cite{zhang2025dreamvla}, Fast-WAM~\cite{yuan2026fastwam}, and ImageWAM~\cite{zhang2026imagewam}. The complete comparison sets are provided in Tables~\ref{tab:libero}--\ref{tab:vlabench}; reproduced results are marked in the corresponding table captions.、
We compare our method with representative VLA and world-model approaches, including ACoT-VLA~\cite{zhong2026acotvla}, Octo~\cite{mees2024octo}, OpenVLA~\cite{kim2024openvla}, $\pi_0$~\cite{black2024pi_0}, $\pi_{0.5}$~\cite{physicalintelligence2025pi05}, CoT-VLA~\cite{zhao2025cotvla}, WorldVLA~\cite{cen2025worldvla}, DreamVLA~\cite{zhang2025dreamvla}, Fast-WAM~\cite{yuan2026fastwam}, and ImageWAM~\cite{zhang2026imagewam}.
The complete comparison sets for each benchmark are reported in Tables~\ref{tab:libero}--\ref{tab:vlabench}, with reproduced results indicated in the corresponding table captions.

%Our primary baseline is ACoT-VLA, which uses explicit and implicit action chain-of-thought reasoning for action generation. We also compared with reported VLA baselines under the same benchmark protocols, including $\pi_0$, $\pi_{0.5}$, and {...} \textcolor{red}{please finish this paragraph}

% \textcolor{red}{please do not use {...}. You can use , etc., as shown in Table XXX}. 
%These baselines appear only in the corresponding result tables, following the evaluation style of ACoT-VLA.

\noindent\textbf{Implementation Details.}
% We build on the PaliGemma-based ACoT-VLA backbone, using SigLIP So400m/14 as the vision encoder, Gemma-2B as the multimodal prefix model, and Gemma-300M modules as action experts. Our latent spatiotemporal reasoning module predicts $K=10$ future SigLIP latent deltas, supervised by latent MSE and cosine direction losses. The optional pixel reconstruction head is used only during training and is removed at inference.
% % 
% For LIBERO, we train for 100K steps with batch size 16, AdamW, gradient clipping 1.0, EMA decay 0.999, and peak learning rate $5\times10^{-6}$. The future-frame offsets are $(15,30,45,60,75,90,105,120,135,150)$, with reconstruction and cosine loss weights both set to 0.1. For the controlled VLABench setting, we follow the released ACoT-VLA hyperparameters: 61K steps, batch size 128, peak learning rate $5\times10^{-5}$, 10K warmup, trainable vision encoder, frozen LLM backbone, trainable token embeder, and $\pi_{0.5}$ initialization. VLABench runs are conducted on 8 NVIDIA H100 GPUs with FSDP sharding.
We use 8 NVIDIA H100 GPUs to train the reasoning module and use 4 NVIDIA A6000 GPUs to train the gist memeory module, while we only use a single NVIDIA A6000 GPU for end-to-end inference latency evaluation.
We build on the PaliGemma-based backbone \cite{zhong2026acotvla}, using SigLIP So400m/14 as the vision encoder, Gemma-2B as the multimodal prefix model, and Gemma-300M modules as action experts.
Latent spatiotemporal reasoning module predicts $K=10$ future SigLIP latent states, supervised by latent MSE and cosine direction losses, while the auxiliary pixel reconstruction head is used only during training.
For LIBERO, we train for 100K steps with a batch size of 16 and a peak learning rate of $5\times10^{-6}$ (cosine decay to $1 \times 10^{-7}$).
For VLABench, we follow the configuration of \cite{zhong2026acotvla} with 61K training steps and a batch size of 128 to ensure a controlled comparison.

\begin{table}[t]
\centering
\scriptsize 
\setlength{\tabcolsep}{4pt}
\renewcommand{\arraystretch}{0.9}

\caption{Comparison on the LIBERO benchmark. * represents that the LLM
backbone is frozen during training.
Our proposed approach is trained on the LIBERO dataset.
All metrics are average success rates (\%).
The best results are highlighted in \textbf{bold}. $\triangle$ denotes IG-VLA with Latent Spatiotemporal Reasoning replaced by Scene Gist Memory.}
\vspace{-0.1in}
\label{tab:libero}

\resizebox{\textwidth}{!}{%
\begin{tabular}{l cc cc cc cc cc}
\toprule

\multirow{2}{*}{Methods} &
\multicolumn{2}{c}{Spatial} &
\multicolumn{2}{c}{Object} &
\multicolumn{2}{c}{Goal} &
\multicolumn{2}{c}{Long} &
\multicolumn{2}{c}{Avg.} \\

\cmidrule(lr){2-3}
\cmidrule(lr){4-5}
\cmidrule(lr){6-7}
\cmidrule(lr){8-9}
\cmidrule(lr){10-11}

& SR $\uparrow$ & Rank $\downarrow$
& SR $\uparrow$ & Rank $\downarrow$
& SR $\uparrow$ & Rank $\downarrow$
& SR $\uparrow$ & Rank $\downarrow$
& SR $\uparrow$ & Rank $\downarrow$ \\

\midrule

Diffusion Policy
& 78.3 & 29 & 92.5 & 21 & 68.3 & 30 & 50.5 & 30 & 72.4 & 30 \\

Octo
& 78.9 & 28 & 85.7 & 29 & 84.6 & 24 & 51.1 & 29 & 75.1 & 28 \\

\midrule

CoT-VLA
& 87.5 & 23 & 91.6 & 23 & 87.6 & 22 & 69.0 & 22 & 81.1 & 23 \\

WorldVLA (8) (256$\times$256)
& 85.6 & 25 & 89.0 & 26 & 82.6 & 26 & 59.0 & 25 & 79.1 & 24 \\

WorldVLA (8) (512$\times$512)
& 87.6 & 22 & 96.2 & 18 & 83.4 & 25 & 60.0 & 24 & 81.8 & 22 \\

DreamVLA
& 97.5 & 11 & 94.0 & 19 & 89.5 & 18 & 89.5 & 15 & 92.6 & 17 \\

UniVLA
& 95.4 & 17 & 98.8 & 7 & 93.6 & 15 & 94.0 & 9 & 95.5 & 13 \\

F1
& 98.2 & 6 & 97.8 & 13 & 95.4 & 14 & 91.3 & 13 & 95.7 & 12 \\

GE-Act
& 98.2 & 6 & 97.6 & 15 & 95.8 & 12 & 94.4 & 8 & 96.5 & 10 \\

\midrule

Fast-WAM
& 98.2 & 6 & 100.0 & 1 & 97.0 & 10 & 95.2 & 5 & 97.6 & 5 \\
ImageWAM
& 97.2 & 12 & 99.2 & 4 & 98.8 & 4 & 98.4 & 1 & 98.4 & 4 \\

\midrule

TraceVLA
& 84.6 & 27 & 85.2 & 30 & 75.1 & 29 & 54.1 & 27 & 74.8 & 29 \\

OpenVLA
& 84.7 & 26 & 88.4 & 27 & 79.2 & 27 & 53.7 & 28 & 76.5 & 27 \\

UniAct
& 77.0 & 30 & 87.0 & 28 & 77.0 & 28 & 70.0 & 21 & 77.8 & 26 \\

SpatialVLA
& 88.2 & 21 & 89.9 & 25 & 88.6 & 20 & 55.5 & 26 & 78.1 & 25 \\

ThinkAct
& 88.3 & 20 & 91.4 & 24 & 87.1 & 23 & 70.9 & 20 & 84.4 & 21 \\

$\pi_0$-FAST
& 96.4 & 15 & 96.8 & 17 & 88.6 & 20 & 60.2 & 23 & 85.5 & 20 \\

FPC-VLA
& 87.0 & 24 & 92.0 & 22 & 89.2 & 19 & 82.2 & 18 & 86.9 & 19 \\

SmoVLA
& 93.0 & 19 & 94.0 & 19 & 91.0 & 17 & 77.0 & 19 & 88.8 & 18 \\

GR00T-N1
& 94.4 & 18 & 97.6 & 15 & 93.0 & 16 & 90.6 & 14 & 93.9 & 16 \\

$\pi_0$
& 96.8 & 14 & 98.8 & 7 & 95.8 & 12 & 85.2 & 17 & 94.1 & 15 \\

GO-1
& 96.2 & 16 & 97.8 & 13 & 96.0 & 11 & 89.2 & 16 & 94.8 & 14 \\

DD-VLA
& 97.2 & 12 & 98.6 & 9 & 97.4 & 8 & 92.0 & 12 & 96.3 & 11 \\

MemoryVLA
& 98.4 & 5 & 98.4 & 10 & 97.4 & 8 & 93.4 & 10 & 96.7 & 9 \\

$\pi_{0.5}$
& 98.8 & 3 & 98.2 & 12 & 98.0 & 6 & 92.4 & 11 & 96.9 & 8 \\

OpenVLA-OFT
& 97.6 & 10 & 98.4 & 10 & 97.9 & 7 & 94.5 & 7 & 97.1 & 7 \\

VLA-Adapter
& 97.8 & 9 & 99.2 & 4 & 99.2 & 2 & 95.0 & 6 & 97.3 & 6 \\

ACoT-VLA*
& 99.4 & 1 & 99.6 & 3 & 98.8 & 4 & 96.0 & 4 & 98.5 & 2 \\

ACoT-VLA*
& 98.6 & 4 & 99.0 & 6 & 99.4 & 1 & 97.0 & 3 & 98.5 & 2 \\

\midrule

\textbf{IG-VLA (Ours)}
& \textbf{99.4} & 1
& \textbf{99.8} & 2
& 99.2 & 2
& \textbf{97.2} & 2
& \textbf{98.9} & 1 \\

\textbf{IG-VLA (Ours)$^{\triangle}$}
& \textbf{99.4} & 1
& \textbf{100.0} & 1
& 98.2 & 5
& \textbf{97.8} & 2
& \textbf{98.9} & 1 \\
\bottomrule
\end{tabular}%
}
\vspace{-0.1in}
\end{table}

\begin{table}[t]
\centering
\scriptsize
\setlength{\tabcolsep}{6pt}
\renewcommand{\arraystretch}{0.9}

\caption{
Supervised fine-tuning results on LIBERO-Plus.
$\diamond$ denotes a pipeline with a frozen LLM backbone, and
$\triangle$ denotes IG-VLA with Latent Spatiotemporal Reasoning replaced by Scene Gist Memory.
Best results are in \textbf{bold}.
}
\vspace{-0.1in}
\label{tab:libero_plus_sft}

\resizebox{\textwidth}{!}{%
\begin{tabular}{l ccccccc c}
\toprule
Methods 
& Camera 
& Robot 
& Language 
& Light 
& Background 
& Noise 
& Layout 
& Avg. \\
\midrule

$\pi_0^{\diamond}$
& 79.6 & 21.1 & 72.5 & 84.7 & 86.2 & 68.3 & 69.4 & 67.4 \\

$\pi_{0.5}^{\diamond}$
& 70.3 & 41.7 & 81.1 & 97.3 & 94.6 & 71.8 & 84.9 & 75.7 \\

ACoT-VLA$^{\diamond}$
& 91.2 & 62.5 & 80.3 & 95.1 & 91.5 & 88.3 & 84.9 & 84.1 \\

ACoT-VLA
& 96.6 & 70.4 & 79.7 & 95.1 & 97.1 & 95.9 & 85.0 & 88.0 \\

\midrule

\textbf{IG-VLA (Ours)}
& 95.4 & 68.9 & \textbf{85.6} & \textbf{97.5} & 97.2 & 94.9 & \textbf{85.5} & \textbf{88.5} \\

\textbf{IG-VLA (Ours)$^{\triangle}$}
& 95.3 & 69.0 & 85.1 & 97.4 & \textbf{97.5} & 94.7 & 84.6 & 88.3 \\

\bottomrule
\end{tabular}%
}

\vspace{-0.2in}
\end{table}

\subsection{Simulation Experiments}

\noindent\textbf{Main Results.}
Tables~\ref{tab:libero}--\ref{tab:vlabench} present the simulation results on LIBERO, LIBERO-Plus, and VLABench.
On LIBERO, both IG-VLA (Reasoning Policy) and IG-VLA$^{\triangle}$ (Gist Policy) achieve the highest average success rate of 98.9\%, outperforming the strongest baseline ACoT-VLA of 98.5\%.
In particular, IG-VLA$^{\triangle}$ achieves 100.0\% and 97.8\% success rates on the Object and Long suites, respectively, demonstrating that Scene Gist preserves strong manipulation performance without explicit future reasoning rollout.
On LIBERO-Plus, our supervised fine-tuning achieves the best average success rate of 88.5\%, exceeding best baseline ACoT-VLA by 0.5 percentage points, with a notable improvement on language variation from 79.7\% to 85.6\%. 
% We also report LIBERO-Plus Zero-Shot Transfer result in Appendix \ref{more}.
Consistent improvements are also observed on VLABench, where our method achieves the highest average Intention Score (IS) of 66.68, improving over ACoT-VLA by 3.18 points.
The improvement is particularly pronounced under semantic-instruction variation, where IS and PS increase from 56.8/39.6 to 65.2/48.49, respectively.
Although the average PS on VLABench is slightly lower than ACoT-VLA (46.11 vs.\ 47.40), the overall results across the three benchmarks demonstrate that latent spatiotemporal reasoning improves intention understanding and action prediction, while Scene Gist effectively internalizes these reasoning benefits for efficient manipulation without explicit future rollout.
% \textbf{LIBERO Benchmark.}
% We first evaluate on the standard LIBERO benchmark. Models are trained on LIBERO demonstrations and evaluated on the corresponding LIBERO task suites, including spatial, object, goal, and long-horizon manipulation tasks. Table~\ref{tab:libero} presents the quantitative results.
% Both our reasoning policy $\pi^{R}$ and gist policy $\pi^{G}$ achieve the best overall performance, reaching an average success rate of 98.9\%, outperforming the strongest baseline ACoT-VLA by 0.4 percentage points.
% The reasoning policy achieves 99.4\%, 99.8\%, and 97.2\% on the Spatial, Object, and Long suites, respectively, improving over ACoT-VLA by up to 0.8 percentage points.
% Notably, the gist policy maintains the same 98.9\% average success rate without explicit future reasoning rollout, while achieving 100.0\% on Object and 97.8\% on Long, surpassing ACoT-VLA by 1.0 and 0.8 percentage points, respectively.
% These results demonstrate that spatiotemporal future reasoning improves action prediction, while the Scene Gist effectively internalizes reasoning-derived behavioral knowledge and preserves strong manipulation performance without explicit future rollout at inference.

\begin{table}[H]
\centering
\scriptsize
\setlength{\tabcolsep}{5pt}
\renewcommand{\arraystretch}{0.9}

\caption{
Comparison on VLABench.
IS and PS denote Intention Score and Progress Score.
$\diamond$ denotes pipeline with a frozen LLM backbone, and $\triangle$ denotes IG-VLA with Latent Spatiotemporal Reasoning replaced by Scene Gist Memory.
Best results are in \textbf{bold}.
}
\vspace{-0.1in}
\label{tab:vlabench}

\resizebox{\textwidth}{!}{%
\begin{tabular}{l c cc cc cc cc cc cc}
\toprule

\multirow{2}{*}{Methods}
& \multirow{2}{*}{Guidance}
& \multicolumn{2}{c}{In-dist.}
& \multicolumn{2}{c}{Category}
& \multicolumn{2}{c}{Commonsense}
& \multicolumn{2}{c}{Instruction}
& \multicolumn{2}{c}{Texture}
& \multicolumn{2}{c}{Avg.} \\

\cmidrule(lr){3-4}
\cmidrule(lr){5-6}
\cmidrule(lr){7-8}
\cmidrule(lr){9-10}
\cmidrule(lr){11-12}
\cmidrule(lr){13-14}

& 
& IS $\uparrow$ & PS $\uparrow$
& IS $\uparrow$ & PS $\uparrow$
& IS $\uparrow$ & PS $\uparrow$
& IS $\uparrow$ & PS $\uparrow$
& IS $\uparrow$ & PS $\uparrow$
& IS $\uparrow$ & PS $\uparrow$ \\

\midrule

$\pi_0^{\diamond}$
& Linguistics
& 67.8 & 62.7
& 44.0 & 33.6
& 54.9 & \textbf{43.0}
& 58.0 & 38.7
& 50.6 & 42.5
& 55.0 & 44.1 \\

$\pi_{0.5}^{\diamond}$
& Linguistics
& 75.0 & 60.8
& 49.6 & 35.3
& 57.5 & 41.6
& 57.1 & 30.3
& 62.0 & 47.4
& 60.2 & 43.1 \\

ACoT-VLA
& \textbf{Action}
& 79.8 & \textbf{66.1}
& 54.1 & \textbf{38.9}
& 52.3 & 37.8
& 56.8 & 39.6
& 74.6 & \textbf{54.6}
& 63.5 & \textbf{47.4} \\

\midrule

\textbf{IG-VLA (Ours)}
& \textbf{Action}
& \textbf{80.80} & 61.64
& 54.20 & 32.21
& \textbf{58.20} & 38.97
& \textbf{65.20} & \textbf{48.49}
& \textbf{74.99} & 49.23
& \textbf{66.68} & 46.11 \\
\textbf{IG-VLA (Ours)$^{\triangle}$}
& \textbf{Action}
& 80.48 & 61.23
& \textbf{54.35} & 31.75
& 57.82 & 39.18
& 64.76 & 48.22
& 74.71 & 48.88
& 66.26 & 45.68 \\

\bottomrule
\end{tabular}%
}
\vspace{-0.2in}
\end{table}

\noindent\textbf{End-to-end Inference Speedup.}
Table~\ref{tab:latency} compares the end-to-end inference latency of our method with existing baselines on LIBERO using a single NVIDIA A6000 GPU. We report latency per action chunk to characterize the computational cost of generating each action sequence. With Scene Gist Memory, IG-VLA$^{\triangle}$ requires only 169.5\,ms per action chunk, achieving $1.78\times$ and $1.55\times$ speedups over FastWAM (1-step) (302\,ms) and ImageWAM (263\,ms), respectively. Alongside this inference efficiency, the gist policy achieves a 98.9\% average success rate on LIBERO, surpassing ImageWAM (98.4\%) and the strongest baseline, ACoT-VLA (98.5\%). These results indicate that the lower inference cost is accompanied by strong manipulation performance. By internalizing reasoning-derived scene--behavior associations into a compact Scene Gist Token, the gist policy guides action prediction without repeatedly performing explicit future rollouts. This allows IG-VLA$^{\triangle}$ to retain the benefits of future reasoning while reducing its inference overhead, supporting efficient VLA deployment. Baseline results are taken from the ImageWAM paper~\citep{zhang2026imagewam}.

% \noindent\textbf{End-to-end Inference Speedup.} 
% Table~\ref{tab:latency} compares end-to-end inference latency of our proposed method with baselines on LIBERO using a single NVIDIA A6000 GPU. IG-VLA$^{\triangle}$ requires only 169.5\,ms per action chunk, achieving $1.78\times$ and $1.55\times$ speedups over FastWAM (1-step) (302\,ms) and ImageWAM (263\,ms), respectively. It also achieves a 98.9\% average success rate on LIBERO, surpassing ImageWAM (98.4\%) and the strongest baseline, ACoT-VLA (98.5\%). These results show that Scene Gist combines efficient inference with strong task performance without explicit future rollouts. Baseline results are taken from ImageWAM~\citep{zhang2026imagewam} paper.

\begin{table}[H]
\vspace{-0.1in}

\centering

% =========================
% Left: Latency
% =========================
\begin{minipage}[t]{0.49\linewidth}
\centering
\small
\setlength{\tabcolsep}{3pt}
\renewcommand{\arraystretch}{0.9}

\caption{Inference latency comparison with baselines using a single NVIDIA A6000 GPU.}
\label{tab:latency}

\vspace{-0.08in}

\resizebox{0.95\linewidth}{!}{
\begin{tabular}{lcc}
\toprule
% Method & Test-time Future Representation & Latency $\downarrow$ \\
\multirow{2}{*}{Method} & Test-time Future & \multirow{2}{*}{Latency $\downarrow$} \\
 & Representation & \\
\midrule
FastWAM-IDM
& Generated video & 1081 ms \\

FastWAM (1-step)
& Cached video features & 302 ms \\

ImageWAM
& Cached image features & 263 ms \\
\midrule

\textbf{IG-VLA (Ours)}
& Latent future rollout & 420.7 ms \\

\textbf{IG-VLA (Ours)$^{\triangle}$}
& Scene gist token & \textbf{169.5 ms} \\
\bottomrule
\end{tabular}
}

\end{minipage}
\hfill
% =========================
% Right: Keyframe Ablation
% =========================
\begin{minipage}[t]{0.49\linewidth}
\centering
\small

\caption{Ablation of different numbers of predicted latent visual keyframes $K$ on LIBERO.}
\label{tab:keyframe_ablation}

% \vspace{-0.08in}

\resizebox{0.99\linewidth}{!}{
\begin{tabular}{lcccc}
\toprule
Suite
& $K=3$
& $K=5$
& $K=10$
& $K=15$ \\
\midrule

LIBERO-Spatial
& 98.4 & 98.6 & \textbf{99.4} & 98.0 \\

LIBERO-Object
& \textbf{99.8} & 99.6 & \textbf{99.8} & 99.2 \\

LIBERO-Goal
& 97.4 & 98.8 & \textbf{99.2} & 98.4 \\

LIBERO-10
& 96.8 & 96.4 & \textbf{97.2} & 95.6 \\
\midrule

Average
& 98.10 & 98.35 & \textbf{98.90} & 97.80 \\
\bottomrule
\end{tabular}
}

\end{minipage}

\vspace{-0.2in}
\end{table}
% Table~\ref{tab:latency} compares the end-to-end inference latency on LIBERO using a single NVIDIA A6000. Our framework with gist token requires only 169.5\,ms per action chunk, faster than FastWAM (302\,ms) and ImageWAM (263\,ms), with speedup of $1.78\times$ and $1.55\times$, respectively. Despite its lower inference cost, the gist policy achieves a 98.9\% average success rate on LIBERO, outperforming ImageWAM (98.4\%) and the strongest baseline ACoT-VLA (98.5\%). These results demonstrate that Scene Gist enables both efficient inference and strong task performance without explicit future rollout. Baseline results are adopted from ImageWAM~\citep{zhang2026imagewam}.
\begin{figure}[t]
\vspace{-0.1in}
    \centering
    \vspace{-0.05in}\includegraphics[width=\linewidth]{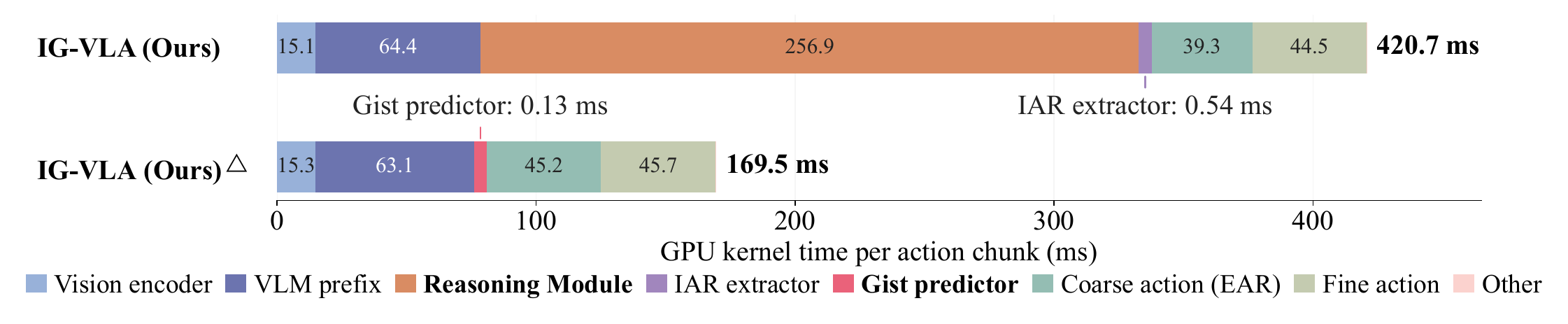}
    \vspace{-0.3in}
    \caption{
    GPU time breakdown(batch size 1, BF16), averaged over three profiled forward passes. 
    % Totals sum GPU kernel durations, excluding host-side and synchronization overhead. 
    % IAR and gist segments are enlarged for visibility.
    }
    \label{fig:gist_latency}
    \vspace{-0.1in}
\end{figure}

% \vspace{-0.1in}
\noindent\textbf{Latency Breakdown Analysis.} Figure~\ref{fig:gist_latency} further analyzes the computational cost of our reasoning and gist policies. Explicit spatiotemporal reasoning incurs 256.9\,ms of GPU time, whereas the Scene Gist predictor introduces only 0.13\,ms of additional overhead.
Consequently, replacing explicit future reasoning with Scene Gist reduces the overall GPU time from 420.7\,ms to 169.5\,ms.
Importantly, this substantial efficiency gain does not compromise task performance: our gist policy achieves the highest average success rate of 98.9\% on LIBERO, outperforming the strongest baseline ACoT-VLA (98.5\%) by 0.4 percentage points.

\subsection{Ablation Study}

% \begin{table}[H]
% \vspace{-0.1in}
% \centering
% \small
% \setlength{\tabcolsep}{6pt}
% \renewcommand{\arraystretch}{1.05}
% \caption{
% Success rates (\%) with different number of predicted latent visual keyframes \(K\) on
% LIBERO. 
% % Results are success rates (\%).
% }
% \label{tab:keyframe_ablation}
% \begin{tabular}{lcccc}
% \toprule
% Suite
% & \(K=3\)
% & \(K=5\)
% & \(K=10\)
% & \(K=15\) \\
% \midrule
% LIBERO-Spatial
% & 98.4 & 98.6 & \textbf{99.4} & 98.0 \\
% LIBERO-Object
% & \textbf{99.8} & 99.6 & \textbf{99.8} & 99.2 \\
% LIBERO-Goal
% & 97.4 & 98.8 & \textbf{99.2} & 98.4 \\
% LIBERO-10
% & 96.8 & 96.4 & \textbf{97.2} & 95.6 \\
% \midrule
% Average
% & 98.10 & 98.35 & \textbf{98.90} & 97.80 \\
% \bottomrule
% \end{tabular}
% \vspace{-0.2in}
% \end{table}

\textbf{Effects of Different Number of Latent Keyframes.}
% We investigate the effect of the number of predicted latent visual
% keyframes \(K\) in Table~\ref{tab:keyframe_ablation}. Predicting
% \(K=10\) keyframes achieves the highest average success rate of 98.90\%,
% with the best performance on all 4-suite. Fewer keyframes may provide insufficient temporal coverage of future scene evolution, whereas increasing \(K\) to 15 introduces redundant or uncertain long-horizon predictions. These results suggest that a moderate number of keyframes provides the best balance between temporal coverage and reliable visual reasoning.
Table~\ref{tab:keyframe_ablation} studies the effect of different number of predicted latent spatiotemporal keyframes $K$ on LIBERO.
Increasing $K$ from 3 to 10 consistently improves the average success rate from 98.10\% to 98.90\%, suggesting that a sufficient imagination horizon provides richer future dynamics for action reasoning.
However, further increasing $K$ to 15 reduces the success rate to 97.80\%.
This degradation can be attributed to longer-horizon future prediction, which can accumulate prediction errors and introduce redundant or unreliable future information~\cite{huang2026live,jin2026beyond}, thereby weakening its guidance for downstream action reasoning.
We therefore use $K=10$ as the default setting in main experiments.

\begin{figure}[H]
\noindent
\begin{minipage}[c]{0.45\textwidth}
    \centering
    \includegraphics[width=\linewidth]
    {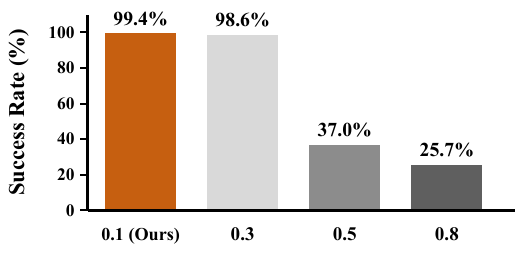}
    \vspace{-0.22in}
    \caption{Ablation of the directional consistency loss weight
    $\lambda_{\mathrm{dir}}$ on LIBERO-Spatial.}
    \label{fig:lambda_dir_ablation}
\end{minipage}
\hfill
\begin{minipage}[c]{0.5\textwidth}
    \centering
    \includegraphics[width=\linewidth]
    {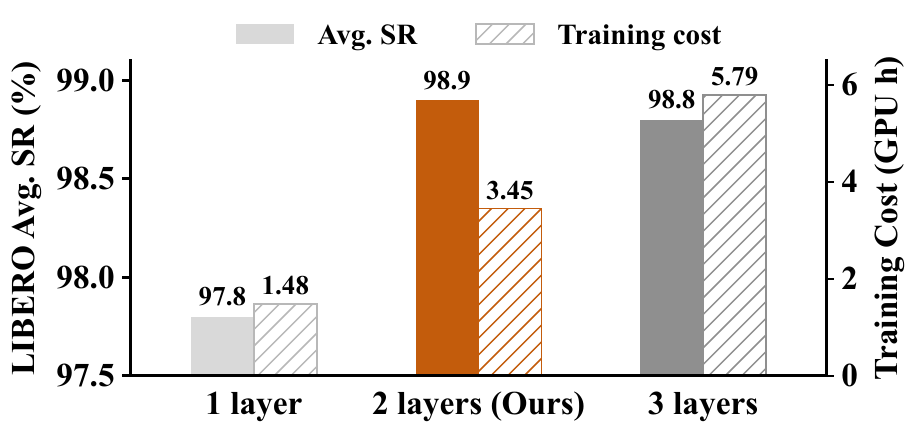}
    \vspace{-0.22in}
    \caption{Effects of different layers of MLP for training gist policy.}
    \label{fig:gist_depth_ablation}
\end{minipage}
\vspace{-0.1in}
\end{figure}

\textbf{Effect of $\lambda_{dir}$.}
We study the relative weighting between the two training objectives by fixing
$\lambda_{\mathrm{reg}}=1$ and varying $\lambda_{\mathrm{dir}}$, thereby controlling the contribution of directional consistency relative to latent transition regression.
As shown in Figure~\ref{fig:lambda_dir_ablation}, $\lambda_{\mathrm{dir}}=0.1$ (main results default setting) achieves the best success rate of 99.4\% on LIBERO-Spatial.
Increasing $\lambda_{\mathrm{dir}}$ to 0.3 slightly reduces the success rate to 98.6\%, while larger weights of 0.5 and 0.8 lead to substantial degradation to 37.0\% and 25.7\%, respectively.
This suggests that the directional objective is most effective as a lightweight auxiliary constraint. A moderate weight encourages alignment with the future visual transition, whereas an overly large weight overemphasizes directional consistency relative to transition-magnitude regression and destabilizes the learning objective.
We therefore set $\lambda_{\mathrm{dir}}=0.1$ and $\lambda_{\mathrm{reg}}=1$ in our main experiments.

\textbf{Effects of Different Number of MLP Layers for Training Gist Policy.}
We ablate the number of MLP layers in the Gist Policy to study the trade-off between model capacity and training efficiency.
As shown in Figure~\ref{fig:gist_depth_ablation}, increasing the depth from one to two layers improves the average LIBERO success rate from 97.8\% to 98.9\%, indicating that a single layer provides insufficient capacity to effectively capture the reasoning-derived Scene Gist.
Further increasing the depth to three layers provides no additional benefit, slightly decreasing the success rate to 98.8\% while increasing the training cost from 3.45 to 5.79 GPU hours.
We therefore adopt a two-layer MLP for the Gist Policy, which provides sufficient capacity while avoiding unnecessary training overhead.

\subsection{Real-World Deployment}
% \textcolor{red}{@Jingwu, put figure 5 and figure 6 in the same row.} 
% \textcolor{red}{(maybe put some details to appendix. Highlight some basic setting and results. maybe we do not need almost one page to introduce real world deployment)}
% \textbf{Setup.} To evaluate our method beyond simulation and on an embodiment different from the training platform, we deploy the fine-tuned policy on a dual-arm UR3 setup (Figure~\ref{fig:ur3_task}). The hardware consists of two 6-DoF UR3 arms with parallel-jaw grippers, yielding a 14-dimensional state/action space, and observations from three RGB cameras (one top-down scene camera and two wrist cameras), each resized to $224\times224$ before the SigLIP encoder. We consider a pick-and-place-a-screwdriver task: starting from the robot home pose, the right arm grasps a screwdriver and places it at a target location (a single-arm task; the left gripper is disabled during deployment). All policies are fine-tuned from \texttt{pi05\_base} on 77 teleoperated demonstrations (15\,Hz).

\begin{wrapfigure}[14]{r}{0.65\textwidth}
    \vspace{-0.2in}
    \includegraphics[width=0.65 \textwidth]{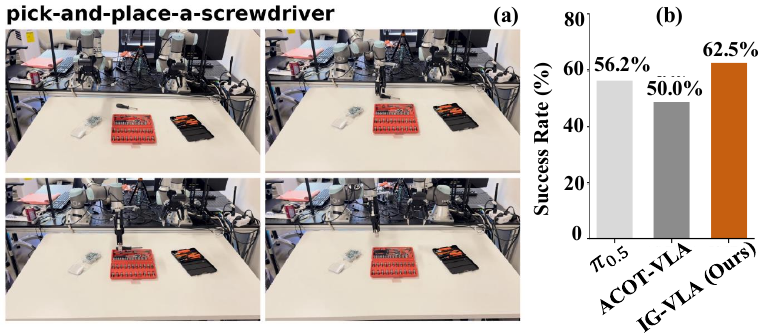}
    \vspace{-0.2in}
    \caption{Real-world evaluation on the dual-arm UR3. (a) Pick-and-place-a-screwdriver task execution.
    % : the right arm approaches, grasps, lifts, and places the screwdriver. 
    (b) Success rate comparison with baselines over 16 trials per method.}
    % \vspace{-0.2in}
    \label{fig:ur3_real}
\end{wrapfigure}

\textbf{Real-World Evaluation.}
% To evaluate our method beyond simulation, 
We further deploy the fine-tuned policy on a dual-arm UR3 robotic platform equipped with parallel-jaw grippers and three RGB cameras, as shown in Figure~\ref{fig:ur3_real}(a).
We consider a pick-and-place task in which the robot grasps a screwdriver and places it at a target location, and fine-tune all policies on 77 teleoperated demonstrations.
Under identical deployment and fine-tuning settings, we compare our method with $\pi_{0.5}$ and ACoT-VLA, conducting 16 trials for each policy with randomized initial screwdriver placements.
As shown in Figure~\ref{fig:ur3_real}(b), our method achieves the highest success rate of 62.5\%, compared with 56.2\% for $\pi_{0.5}$ and 50.0\% for ACoT-VLA.
These results further demonstrate that incorporating future spatiotemporal reasoning improves manipulation performance beyond the underlying VLA backbone and prior action-reasoning approaches.
Detailed deployment settings and action-space conversion are provided in Appendix \ref{real}.

\section{Conclusion}

In this work, we present IG-VLA to address two key limitations of existing VLA reasoning: the lack of explicit anticipation of future scene dynamics and the high inference cost of repeated reasoning rollouts.
By reasoning over future dynamics in latent visual space, IG-VLA provides task-relevant spatiotemporal guidance for action generation, while Scene Gist Memory further converts the acquired reasoning knowledge into compact behavioral guidance for efficient inference.
Experiments across LIBERO, LIBERO-Plus, and VLABench consistently demonstrate improved manipulation performance and generalization, with particularly strong gains under challenging language and semantic variations.
Meanwhile, the gist policy retains competitive reasoning performance while substantially reducing inference cost, showing that explicit future reasoning need not be repeatedly executed at deployment.
Real-world robotic experiments further validate the effectiveness of the proposed framework beyond simulation.
Overall, our results suggest that anticipating future scene dynamics and subsequently internalizing such reasoning provides an effective direction toward more capable and efficient VLA policies.

\section*{AI use statement}

In this work, large language models (LLMs) were used solely as a general-purpose writing assistant. Their role was limited to correcting grammar, fixing typographical errors, and polishing the language for clarity and readability.

% \textcolor{red}{please add}

% \textcolor{red}{please add}

% \section*{Limitations}

% TODO
% \section*{Ethics Statement}

% TODO

\bibliography{iclr2027_conference}
\bibliographystyle{iclr2027_conference}

\newpage

\appendix
\section{Appendix}

\subsection{Overall Detailed Pipeline}

\begin{algorithm}
\caption{Learning and Inference with Latent Spatiotemporal Reasoning and Scene Gist}
\label{alg:scene_gist}
\begin{algorithmic}[1]

\REQUIRE Demonstration dataset $\mathcal{D}$, vision encoder $E_v$,
vision-language backbone $\mathcal{V}$,
spatiotemporal reasoning module $\mathcal{R}_{\phi}$,
reasoning policy $\pi^R_{\theta}$,
gist compressor $\mathcal{C}_{\eta}$,
gist encoder $\mathcal{G}_{\psi}$,
gist policy $\pi^G_{\theta}$,
reasoning horizon $H$

\STATE \textbf{Training}

\FOR{each training sample
$(o_t,L_t,\{o_{t+k},a_{t+k}\}_{k=1}^{H})\sim\mathcal{D}$}

    \STATE $z_t \leftarrow E_v(o_t),\quad
    h_t \leftarrow \mathcal{V}(z_t,L_t)$
    \hfill $\triangleright$ Encode current observation

    \FOR{$k=1,\ldots,H$}

        \STATE $z_{t+k}\leftarrow E_v(o_{t+k})$
        \hfill $\triangleright$ Encode ground-truth future observation

        \STATE $z^G_{t+k}\leftarrow z_{t+k}-z_t$
        \hfill $\triangleright$ Construct ground-truth visual transition

        \STATE $z^F_{t+k}
        \leftarrow \mathcal{R}_{\phi}(h_t,L_t,k)$
        \hfill $\triangleright$ Predict future visual transition

        \STATE $\hat z_{t+k}
        \leftarrow z_t+z^F_{t+k}$
        \hfill $\triangleright$ Recover anticipated future representation

        \STATE $a^R_{t+k}
        \leftarrow
        \pi^R_{\theta}(h_t,L_t,\hat z_{t+k})$
        \hfill $\triangleright$ Predict reasoning-informed action

    \ENDFOR

    \STATE $\displaystyle
    \mathcal{L}_{\mathrm{reg}}
    =
    \frac{1}{H}
    \sum_{k=1}^{H}
    \left\|
    z^F_{t+k}
    -
    z^G_{t+k}
    \right\|_2^2$
    \hfill $\triangleright$ Visual transition regression

    \STATE $\displaystyle
    \mathcal{L}_{\mathrm{dir}}
    =
    \frac{1}{H}
    \sum_{k=1}^{H}
    \left(
    1-
    \cos(
    z^F_{t+k},
    z^G_{t+k})
    \right)$
    \hfill $\triangleright$ Directional consistency

    \STATE $h_t^R\leftarrow
    \mathrm{IAR}
    \left(h_t,\{\hat z_{t+k}\}_{k=1}^{H}\right)$
    \hfill $\triangleright$ Obtain reasoning-enhanced representation

    \STATE $\bar g_t\leftarrow\mathcal{C}_{\eta}(h_t^R)$
    \hfill $\triangleright$ Construct reasoning-derived Scene Gist target

    \STATE $\bar a^G_{t:t+H-1}
    \leftarrow
    \pi^G_{\theta}(h_t,L_t,\bar g_t)$
    \hfill $\triangleright$ Predict actions with Scene Gist target

    \STATE $\displaystyle
    \mathcal{L}_{\mathrm{comp}}
    =
    \frac{1}{H}
    \sum_{k=0}^{H-1}
    \left\|
    \bar a^G_{t+k}
    -
    \operatorname{sg}(a^R_{t+k})
    \right\|_2^2$
    \hfill $\triangleright$ Train gist compressor

    \STATE $g_t\leftarrow\mathcal{G}_{\psi}(h_t)$
    \hfill $\triangleright$ Predict Scene Gist from current context

    \STATE $a^G_{t:t+H-1}
    \leftarrow
    \pi^G_{\theta}(h_t,L_t,g_t)$
    \hfill $\triangleright$ Predict gist-conditioned action chunk

    \STATE $\displaystyle
    \mathcal{L}_{\mathrm{align}}
    =
    \left\|
    g_t-\operatorname{sg}(\bar g_t)
    \right\|_2^2$
    \hfill $\triangleright$ Align predicted and target Scene Gists

    \STATE $\displaystyle
    \mathcal{L}_{\mathrm{act}}
    =
    \frac{1}{H}
    \sum_{k=0}^{H-1}
    \left\|
    a^G_{t+k}
    -
    \operatorname{sg}(a^R_{t+k})
    \right\|_2^2$
    \hfill $\triangleright$ Align reasoning-policy actions

    \STATE $\mathcal{L}_{\mathrm{gist}}
=
\mathcal{L}_{\mathrm{act}}
+
\lambda_{\mathrm{gist}}
\mathcal{L}_{\mathrm{align}}$
    \hfill $\triangleright$ Train Scene Gist encoder

\ENDFOR

\STATE \textbf{Inference}

\STATE $z_t\leftarrow E_v(o_t),\quad
h_t\leftarrow\mathcal{V}(z_t,L_t)$
\hfill $\triangleright$ Encode current observation and instruction

\STATE $g_t\leftarrow\mathcal{G}_{\psi}(h_t)$
\hfill $\triangleright$ Retrieve Scene Gist

\STATE $a^G_{t:t+H-1}
\leftarrow
\pi^G_{\theta}(h_t,L_t,g_t)$
\hfill $\triangleright$ Predict action chunk without future rollout

\RETURN $a^G_{t:t+H-1}$

\end{algorithmic}
\end{algorithm}

Algorithm~\ref{alg:scene_gist} summarizes the overall training and inference pipeline of IG-VLA.
During training, we first encode the current and future observations into the latent visual space and construct their latent transitions as supervision.
The Latent Spatiotemporal Reasoning module learns to predict these future transitions from the current multimodal context, enabling the reasoning policy to anticipate task-relevant scene evolution and use the resulting future information for action prediction.

We then internalize the acquired future reasoning into Scene Gist Memory.
Specifically, the reasoning-enhanced representation is compressed into a compact Scene Gist target, whose associated actions are aligned with those produced by the reasoning policy.
A lightweight gist encoder subsequently learns to predict this reasoning-derived Scene Gist directly from the current multimodal context, with both representation- and action-level supervision.
In this way, the Scene Gist captures the behavioral knowledge distilled from explicit future reasoning without requiring the future rollout itself.

At inference time, only the current observation and instruction are required.
The gist encoder directly retrieves the corresponding Scene Gist from the current context, which conditions the gist policy to generate the action chunk.
Therefore, the expensive Latent Spatiotemporal Reasoning module and future-state rollout are completely bypassed during deployment, while their reasoning benefits are retained through the learned Scene Gist.

\subsection{More Details about Real-World Deployment}
\label{real}

\textbf{Setup.} To evaluate our method beyond simulation and on an embodiment different from the training platform, we deploy the fine-tuned policy on a dual-arm UR3 setup (Figure~\ref{fig:ur3_real}). The hardware consists of two 6-DoF UR3 arms with parallel-jaw grippers, yielding a 14-dimensional state/action space, and observations from three RGB cameras (one top-down scene camera and two wrist cameras), each resized to $224\times224$ before the SigLIP encoder. We consider a pick-and-place-a-screwdriver task: starting from the robot home pose, the right arm grasps a screwdriver and places it at a target location (a single-arm task; the left gripper is disabled during deployment). All policies are fine-tuned from \texttt{pi05\_base} on 77 teleoperated demonstrations (15\,Hz).

\textbf{Deployment.} Since the UR3 is joint-controlled while the policy predicts actions in a delta end-effector (delta-EEF) space, we integrate each predicted delta-EEF chunk onto the current end-effector pose (obtained by forward kinematics from the measured joints) and map it back to joint targets with a numerical damped-least-squares inverse-kinematics solver (warm-started from the current joints); the robot then executes the resulting joint chunk under real-time joint-position control. This conversion is performed entirely server-side, so the joint-space robot client is unchanged. Actions are executed with a near-synchronous chunking scheme: each 10-step delta-EEF chunk is executed in full at 5\,Hz before the next inference, with a joint-velocity cap of $2.5$\,rad/s.

\textbf{Comparison and protocol.} Under identical deployment conditions we compare three policies: $\pi_{0.5}$, ACoT-VLA (the prior baseline), and our method (with latent spatiotemporal reasoning internalized through Scene Gist Memory). All three share the same fine-tuning configuration (delta-EEF action space, action horizon 10, \texttt{pi05\_base} initialization, batch size 16, constant learning rate $5\times10^{-5}$) and differ only in the method itself. For each policy we run 16 trials with randomized initial screwdriver placements drawn from the same distribution across methods, and report the binary task-success rate.

% \subsection{More Experiments}
% \label{more}

% \begin{table}[t]
% \centering
% \scriptsize
% \setlength{\tabcolsep}{6pt}
% \renewcommand{\arraystretch}{0.9}

% \caption{
% Zero-shot transfer results on LIBERO-Plus.
% * denotes reproduced results, $\diamond$ denotes a pipeline with a frozen LLM backbone, and
% $\triangle$ denotes IG-VLA with Latent Spatiotemporal Reasoning replaced by Scene Gist Memory.
% Best results are in \textbf{bold}.
% }
% \vspace{-0.1in}
% \label{tab:libero_plus_zeroshot}

% \resizebox{\textwidth}{!}{%
% \begin{tabular}{l ccccccc c}
% \toprule
% Methods 
% & Camera 
% & Robot 
% & Language 
% & Light 
% & Background 
% & Noise 
% & Layout 
% & Avg. \\
% \midrule

% WorldVLA
% & 0.1 & 27.9 & 41.6 & 43.7 & 17.1 & 10.9 & 38.0 & 25.0 \\

% OpenVLA
% & 0.8 & 3.5 & 23.0 & 8.1 & 34.8 & 15.2 & 28.5 & 15.6 \\

% NORA
% & 2.2 & 37.0 & 65.1 & 45.7 & 58.6 & 12.8 & 62.1 & 39.0 \\

% UniVLA
% & 1.8 & 46.2 & 69.6 & 69.0 & 81.0 & 21.2 & 31.9 & 42.9 \\

% $\pi_0$-Fast
% & 65.1 & 21.6 & 61.0 & 73.2 & 73.2 & 74.4 & 68.8 & 61.6 \\

% RIPT-VLA
% & 55.2 & 31.2 & 77.6 & 88.4 & 91.6 & 73.5 & 74.2 & 68.4 \\

% OpenVLA-OFT
% & 56.4 & 31.9 & 79.5 & 88.7 & 93.3 & 75.8 & 74.2 & 69.6 \\

% $\pi_0^{*}$
% & 61.0 & 40.8 & 63.5 & 89.3 & 84.1 & 80.1 & 76.4 & 69.4 \\

% $\pi_{0.5}^{*}$
% & 75.8 & 79.4 & 83.3 & 95.5 & 95.0 & 89.6 & 87.0 & 85.7 \\

% ACoT-VLA$^{\diamond}$
% & 68.9 & 80.3 & 84.1 & 95.6 & 93.1 & 81.5 & 88.3 & 83.6 \\

% ACoT-VLA
% & 72.6 & 82.6 & 87.5 & 97.7 & 96.5 & 87.8 & 88.1 & 86.6 \\

% \midrule

% \textbf{IG-VLA (Ours)}
% &  &  &  &  &  &  &  &  \\

% \bottomrule
% \end{tabular}%
% }

% \vspace{-0.1in}
% \end{table}

\end{document}